\documentclass[letterpaper, 10 pt, conference]{ieeeconf}  

\IEEEoverridecommandlockouts                             

\usepackage{graphicx}
\usepackage{multirow}
\usepackage{amssymb}
\usepackage{pifont}
\usepackage{amsmath}
\usepackage[table]{xcolor}

\definecolor{limingcolor}{RGB}{0,102,204}      
\definecolor{alexcolor}{RGB}{180,0,0}          
\definecolor{mariacolor}{RGB}{0,150,90}        
\definecolor{reviewcolor}{RGB}{150,0,150}      

\usepackage[normalem]{ulem}

\usepackage[colorlinks=true,
            linkcolor=blue,
            citecolor=blue,
            urlcolor=blue]{hyperref}
\usepackage{cleveref}

\title{Spotlighting Task-Relevant Features: Object-Centric Representations for Better Generalization in Robotic Manipulation}

\author{Alexandre Chapin$^{1}$, Bruno Machado$^{1}$, Emmanuel Dellandrea$^{1}$, Liming Chen$^{1}$
\thanks{$^{1}$Ecole Centrale de Lyon, CNRS, Universite Claude Bernard Lyon 1, INSA Lyon, Université Lumière Lyon 2, LIRIS, UMR5205, 69130 Ecully, France}
\thanks{Correspondance to :
       {\tt\small alexandre.chapin@ec-lyon.fr}}%
}
\begin{document}

\maketitle

\
\begin{abstract}
Training robotic policies that reliably generalize to novel environments remains a persistent challenge. While state-of-the-art models leverage powerful global or dense visual features, these representations struggle to separate critical task-specific signals from background noise, causing failures under visual shifts. In this work, we investigate \textit{Slot-based Object-Centric Representations (SOCRs)} as a structural solution that decomposes scenes into discrete, actionable entities without supervision. Through a large-scale diagnostic study across simulated (\textsc{MetaWorld}, \textsc{LIBERO}) and real-world tasks, we systematically evaluate SOCRs against dominant baselines. We find that structured abstraction drives inherent robustness: SOCRs drastically outperform standard models under severe lighting, texture, and clutter shifts without task-specific fine-tuning, and they scale effectively with in-domain video pre-training. However, we also identify a critical vulnerability: a structural capacity trade-off that leads to \textit{slot merging} under high clutter. By mapping out both the strengths and the bottleneck limits of SOCRs, this work provides a clear roadmap for integrating structured visual abstractions into next-generation, generalizable robotic systems. Project website: \url{http://spotlight-iros.github.io/}
\end{abstract}

\begin{figure*}[h]
	\centering
	\includegraphics[width=0.85\textwidth]{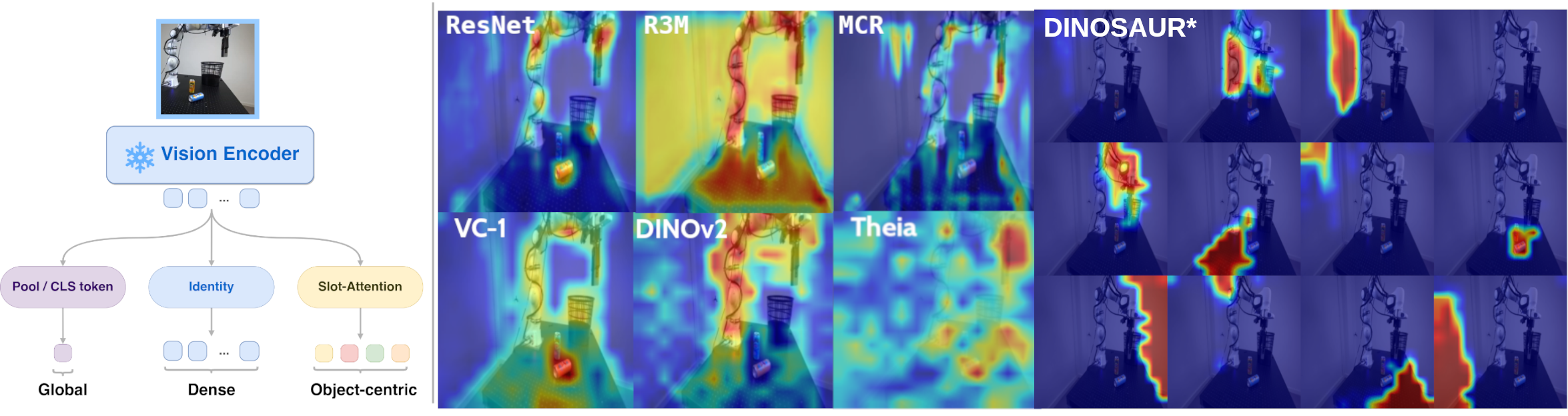}
	\caption{\textbf{Overview of visual representations.} (\textit{Left}) We use a set of pre-trained visual models with different latent-space structures: global, dense, and object-centric. Dense representations are extracted from one of the encoder's layers (CNN or ViT) before linear projection, while global representations are obtained after pooling operations. Slot-based object-centric representations emerge from an additional Slot-Attention layer that binds every dense feature to a finite set of slots. (\textit{Right}) We visualize how each representation attends to different parts of the image (using GradCAM~\cite{Selvaraju_2019} for CNN-based methods and Attention-Rollout~\cite{abnar2020quantifyingattentionflowtransformers} for ViT-based). Most methods focus narrowly and may be distracted by irrelevant regions. In contrast, object-centric representations (DINOSAUR*) attend to multiple parts, naturally separating task-relevant from irrelevant information.}
	\label{fig:overview}
\end{figure*}
 
\section{Introduction}
A \textbf{visuomotor policy} tells a robot what to do based on what it sees given a goal, mapping raw visual inputs to motor actions. State-of-the-art approaches for visuomotor policy learning are data-driven through deep learning, \textit{e.g.}, imitation learning (IL) \cite{li2025roboticmanipulationimitationlearning}, and leverage a few expert demonstrations without the need for explicit programming. As such, the visual representation of the robot visual input is a key to the robot's \textbf{generalization} capability when the learned policy must adapt to objects, environments, and tasks that differ from those encountered during training~\cite{li2025roboticmanipulationimitationlearning}.

Recent work has intensified efforts to improve visual representation learning and their pre-training methodology as the foundation for robust robot policy learning. Advances include large-scale pre-training on egocentric video with time-contrastive and language-aligned objectives~\cite{nair2022r3muniversalvisualrepresentation,majumdar2024searchartificialvisualcortex, ma2023vipuniversalvisualreward}, distilling vision foundation models into compact encoders~\cite{shang2024theiadistillingdiversevision}, and adopting self-supervised schemes such as masked autoencoding~\cite{radosavovic2022realworldrobotlearningmasked}. However, despite their diversity, these methods mostly produce one of two features types: \textbf{global} or \textbf{dense features}. \textbf{Global features} is a single vector summarizing the entire image, typically obtained by pooling operations (\textit{e.g.}, max or average pooling in CNNs) or by extracting a special CLS token in ViTs while \textbf{dense features} is an embedding from one of the encoder's layers (usually the last). While effective for in-distribution tasks, these representations lack an explicit mechanism to separate task-relevant objects from irrelevant background noise, as illustrated in Figure~\ref{fig:overview}. Consequently, the resulting policies often overfit to spurious correlations, such as table texture or lighting conditions, leading to catastrophic failure under distribution shifts~\cite{burns2023makespretrainedvisualrepresentations, hu2023pretrainedvisionmodelsmotor}.

To address these limitations, we revisit a fundamental question in visuomotor control: 
\emph{What structural properties must a visual representation satisfy in order to yield policies that generalize under distribution shift?}

We posit a \textbf{structural bottleneck hypothesis}: representations that explicitly decompose scenes into discrete, object-level entities impose an inductive bias that promotes invariance to low-level appearance changes and reduces reliance on spurious correlations. In contrast, global or dense feature encodings entangle task-relevant and irrelevant signals, forcing downstream policies to implicitly learn separation from mixed representations. Slot-based Object-Centric Representations (SOCRs) ~\cite{locatello2020objectcentriclearningslotattention, seitzer2023bridginggaprealworldobjectcentric} instantiate such a bottleneck through competitive attention over a fixed set of latent slots much in line with theories of human perception~\cite{spelke1990principles, lake2016buildingmachineslearnthink}. While prior work has explored slot-based models in unsupervised perception or simplified synthetic control settings ~\cite{yoon2023investigationpretrainingobjectcentricrepresentations, heravi2023visuomotorcontrolmultiobjectscenes, haramati2024entitycentricreinforcementlearningobject, watters2019cobradataefficientmodelbasedrl, kipf2022conditionalobjectcentriclearningvideo}, these studies have primarily emphasized \textbf{relational complexity}, focusing on multi-object reasoning and physical interactions within visually simplified environments. Consequently, the axis of \textbf{perceptual complexity} remains largely unexplored. It remains unclear whether (i) structured bottlenecks scale to the visual noise and distractors inherent to real-world robotic manipulation, (ii) they provide measurable robustness gains over modern foundation models (\textit{e.g.}, DINOv2~\cite{oquab2024dinov2learningrobustvisual}, Theia~\cite{shang2024theiadistillingdiversevision}), and (iii) what structural limitations govern their behavior under increasing scene complexity.

In this work, we move beyond performance benchmarking and provide the first systematic diagnostic study of SOCRs for robot policy learning. We analyze how representation structure impacts generalization, identify concrete failure modes arising from capacity constraints, and provide guidance on how to overcome problems grounded in this structural understanding.

Our contributions are fourfold:
\begin{itemize}
    \item \textbf{Validating structural robustness.} We empirically validate that SOCRs inherently improve robustness against different distribution shifts across simulated and real-world manipulation tasks.
    
    \item \textbf{Benefits of large-scale pre-training.} We demonstrate that, contrary to existing assumptions~\cite{didolkar2024zeroshotobjectcentricrepresentationlearning}, object-centric methods can effectively leverage large-scale pre-training to significantly boost downstream performance.

    \item \textbf{Identifying key failure modes.} We provide the first systematic breakdown of why object-centric representations fail in downstream control, pinpointing slot merging and capacity limits as the primary bottlenecks.

    \item \textbf{Balancing capacity and robustness.} Through targeted slot-number ablations, we reveal that structural capacity directly dictates out-of-distribution (OOD) robustness.

\end{itemize}

\section{Related works}

\paragraph{Pretrained vision-based models for robot learning} 
Large-scale vision models (MoCo~\cite{chen2020improvedbaselinesmomentumcontrastive}, DINO~\cite{caron2021emergingpropertiesselfsupervisedvision}, DINOv2~\cite{oquab2024dinov2learningrobustvisual}, CLIP~\cite{radford2021learningtransferablevisualmodels}) have proven surprisingly effective for visuomotor policy learning~\cite{parisi2022unsurprisingeffectivenesspretrainedvision}. To improve alignment with manipulation tasks, subsequent works introduced domain-specific pretraining using egocentric or robotic videos (R3M~\cite{nair2022r3muniversalvisualrepresentation, grauman2022ego4dworld3000hours}, VC-1~\cite{majumdar2024searchartificialvisualcortex, russakovsky2015imagenetlargescalevisual}, Theia~\cite{shang2024theiadistillingdiversevision}, MCR~\cite{jiang2024robots}). However, recent studies demonstrate that dataset quality, diversity, and emergent object-segmentation properties often outweigh strict domain alignment for generalization~\cite{burns2023makespretrainedvisualrepresentations, dasari2023unbiasedlookdatasetsvisuomotor}. These findings motivate our shift toward structured, object-centric representations that naturally capture task-relevant scene composition.

\paragraph{Slot-based object-centric representations} 
Slot-based object-centric representations (SOCRs) decompose visual scenes into structured latent entities (slots) in an unsupervised manner, gaining traction in autonomous driving, robotics, and explainability~\cite{heravi2023visuomotorcontrolmultiobjectscenes, hamdan2024carformerselfdrivinglearnedobjectcentric, mosbach2025soldslotobjectcentriclatent, wang2024explainableimagerecognitionenhanced}. Originating from generative models~\cite{watters2019cobradataefficientmodelbasedrl, kabra2021simoneviewinvarianttemporallyabstractedobject, burgess2019monetunsupervisedscenedecomposition}, methods like Slot Attention~\cite{locatello2020objectcentriclearningslotattention} have evolved to incorporate advanced decoders~\cite{jiang2023objectcentricslotdiffusion, wu2023slotdiffusionobjectcentricgenerativemodeling, singh2022illiteratedallelearnscompose}, pretrained backbones~\cite{seitzer2023bridginggaprealworldobjectcentric}, and temporal dynamics for video~\cite{kipf2022conditionalobjectcentriclearningvideo, elsayed2022saviendtoendobjectcentriclearning,singh2022simpleunsupervisedobjectcentriclearning, zadaianchuk2023objectcentriclearningrealworldvideos}. While SOCRs show potential for aiding generalization~\cite{burns2023makespretrainedvisualrepresentations, jiang2024robots}, their application in control or imitation learning remains largely limited to simple, synthetic environments and are evaluated on in-domain scenarios only~\cite{yoon2023investigationpretrainingobjectcentricrepresentations, heravi2023visuomotorcontrolmultiobjectscenes, haramati2024entitycentricreinforcementlearningobject}. Crucially, these existing approaches fail to assess how well SOCRs generalize to new distribution scenarios, such as visual distractors, novel object configurations, or altered backgrounds. Without rigorous out-of-distribution (OOD) testing, it remains unclear whether the theoretical compositionality of slots translates into robust, real-world control policies under environmental shifts. To address this critical gap, our work systematically evaluates SOCRs in realistic robotic manipulation tasks under distributional shifts, comparing them against state-of-the-art dense visual representations.

\paragraph{Segmentation-driven object decomposition for robotics}
An alternative approach pairs supervised foundation models like SAM~\cite{kirillov2023segment} with frozen vision backbones to generate object masks (e.g., POCR~\cite{shi2024composingpretrainedobjectcentricrepresentations}, GROOT~\cite{zhu2023learninggeneralizablemanipulationpolicies}, HODOR~\cite{qian2024taskorientedhierarchicalobjectdecomposition}). While effective, these methods face key limitations: they rely on heavily annotated pretraining datasets, are memory and compute intensive (hindering real-time application), and often require explicit spatial prompts like bounding boxes. Conversely, SOCRs learn object decomposition end-to-end without supervision, offering a more flexible and computationally efficient alternative for real-time robotic inference.

\section{Method}
\label{sec:method}

We first introduce the process of extracting object-centric representations from dense features. Then, we detail the integration of the object-centric visual features into a policy training framework.

\paragraph{Object-centric representation} Given an input image $I$, the goal is to produce a set of $K$ object representations, or slots, $S = {s_1, \ldots, s_K}$. To achieve this, the image is first encoded by a visual backbone into dense feature tokens $F = {f_1, \ldots, f_N}$, where $N \gg K$. Slot Attention \cite{locatello2020objectcentriclearningslotattention} then extracts a compact set of object representations $S$ by iteratively attending to these features. It is a differentiable module that performs iterative cross-attention with competition, encouraging different slots to specialize in distinct parts of the input image. Formally, attention weights are computed as:
\begin{equation}
    \mathbf{A} = \mathrm{softmax}\!\left(\frac{\mathbf{Q}\mathbf{K}^T}{\sqrt{D}}\right),
    \quad
    S^{(i+1)} = \mathbf{A}\mathbf{V},
\end{equation}

where the queries $\mathbf{Q}$ are projections of the current slot representations $S^{(i)}$, and the keys $\mathbf{K}$ and values $\mathbf{V}$ are projections of the feature tokens $F$. $D$ is the projected feature dimension. This iterative refinement process yields the final object-centric slots $S$.

The SOCR model we used, which we refer to as DINOSAUR*, builds upon DINOSAUR~\cite{seitzer2023bridginggaprealworldobjectcentric} while introducing \textbf{two key modifications} to enhance representation quality and temporal consistency. First, we modernize the visual backbone by replacing the original DINO \cite{caron2021emergingpropertiesselfsupervisedvision} encoder with the more robust DINOv2 \cite{oquab2024dinov2learningrobustvisual}. Like the original model, we utilize a decoder to reconstruct these high-level backbone features rather than raw pixels. Second, we extend the static Slot-Attention mechanism to the temporal domain. Inspired by \cite{elsayed2022saviendtoendobjectcentriclearning, zadaianchuk2023objectcentriclearningrealworldvideos}, we incorporate a Transformer layer between timesteps to facilitate recursive information transfer. In this setup, slots at each timestep first undergo slot-wise self-attention within the Transformer layer to aggregate context; these refined slots then serve as the initialization for the subsequent Slot-Attention module.

\paragraph{Policy training}
Given a dataset of expert demonstrations $\mathcal{D} = {\tau_1, \ldots, \tau_n}$, where each trajectory $\tau_i = [(o_0, a_0), \ldots, (o_T, a_T)]$ pairs observations with actions, we aim to learn a policy $\pi$ that learns to map the current observation $o_t$ (e.g. visual inputs) to the next action $a_t$. To ensure a fair comparison across different feature types, we utilize a framework inspired by~\cite{haldar2024bakuefficienttransformermultitask}. It consists of an encoder, an observation trunk, and a policy head. The observation trunk is a transformer encoder that encodes a sequence of $T$ past observations, each containing visual features, proprioceptive states, and language embeddings, interleaved with learnable action tokens. The policy head, a MLP, generates the next action from the action tokens. The architecture treats visual inputs as sets of tokens, this property ensures that the transformer can attend over both structured (slot-based) and unstructured (global or dense) features, thereby preserving fairness across representation types . 
Importantly, each policy is trained in a \textbf{multi-task setting}, enabling learning from demonstrations across multiple tasks rather than a single one. All pretrained vision encoders are \textbf{kept frozen} during policy training to isolate the effect of representation choice.

\paragraph{Robotic pre-training}
While object-centric models excel at structured scene decomposition, they are primarily trained on \textit{in-the-wild} datasets. This can result in distribution shift when these models are applied to robotic manipulation environments. To bridge this gap and enhance data diversity, aligning with the observations in~\cite{dasari2023unbiasedlookdatasetsvisuomotor}, we introduce a dedicated domain-specific pretraining stage using large-scale robotic video data. To rigorously evaluate the impact of this domain alignment, we develop two versions of our model: \textbf{DINOSAUR*}, trained on the standard COCO dataset for fair comparison against existing baselines, and \textbf{DINOSAUR-Rob*}, which is pre-trained specifically on robotic data. Our robotic pre-training corpus is a weighted combination of real-world robotic datasets~\cite{khazatsky2024droidlargescaleinthewildrobot, walke2024bridgedatav2datasetrobot, brohan2023rt1roboticstransformerrealworld} spanning a vast array of manipulation skills, environments, and robotic embodiments.

\section{Benchmarks and Experimental setup}
\label{sec:benchmarks}

\paragraph{Environments and tasks}
To comprehensively evaluate visual representations for robot manipulation, we selected \textbf{three environments}, two in simulation and one in the real world, chosen for their diversity in task complexity, embodiment, and visual structure.

In simulation, we use MetaWorld \cite{yu2021metaworldbenchmarkevaluationmultitask}, a widely adopted benchmark consisting of tabletop manipulation tasks with a Sawyer robotic arm. MetaWorld provides a controlled, standardized setting, making it ideal for evaluating representations in simple, single-object scenarios and testing basic generalization (distractors, texture, and lighting). We use the same task set as in~\cite{jiang2024robots}. 

To assess multi-object reasoning, we also include LIBERO \cite{liu2023libero}, a recent benchmark featuring scenes across kitchens, offices, and living rooms. We train and evaluate on the LIBERO-90 tasks, which involve multiple objects with diverse appearances and affordances, emphasizing combinatorial generalization and reasoning about object interactions. However, LIBERO's benchmark does not introduce any distributional shifts for evaluation. So we only report results on the training distribution.

\begin{figure*}[h]
	\centering
	\includegraphics[width=0.85\linewidth]{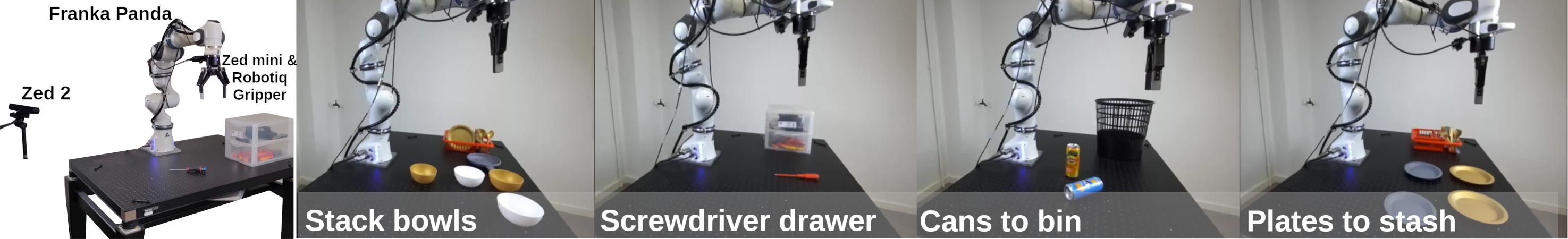}
	\caption{\textbf{Overview of real-world setup.} We evaluate the different visual models on a Franka robotic arm on four tabletop manipulation tasks (From left to right): Stacking bowls into a pan, Opening a drawer placing a screwdriver inside and closing the drawer, Putting cans into a bin and Placing plates into a dish rack.}
	\label{fig:tasks_real}
\end{figure*}
For real-world evaluation, we deploy a Franka robotic arm on a set of four tabletop manipulation tasks, Figure~\ref{fig:tasks_real} provides an overview of the setup used. For each task, we collected 50 expert demonstrations via teleoperation. Task success is computed using success rate. A trial is a success if the final physical configuration meets the task-specific criteria: the bowls are nested inside the pan, the screwdriver is contained within the closed drawer, all cans are positioned inside the bin, or the plates are inside the slots of the dish rack.

\paragraph{Pre-trained visual models} We selected seven representative pre-trained visual encoders, detailed in Table~\ref{tab:models}. This selection spans a diverse architectural landscape, including ResNet variants \cite{he2015deepresiduallearningimage}, Vision Transformers (ViTs) \cite{dosovitskiy2021imageworth16x16words}, and encompasses a variety of training objectives such as supervised learning, self-supervised learning, contrastive learning, and distillation. While these models utilize different pre-training datasets, we follow the findings of Parisi et al. \cite{parisi2022unsurprisingeffectivenesspretrainedvision}, which suggest that the choice of pre-training data is often secondary to the representation architecture for downstream control tasks. Our benchmark aims to capture all major feature representation types to facilitate a comprehensive evaluation: \textbf{global} features (a single token representing the entire image), \textbf{dense} features (a grid of tokens capturing spatial detail), and \textbf{slot-based} features (object-centric representations that decompose the scene into discrete entities).

For transparency regarding computational complexity, Table~\ref{tab:models} reports the total number of tokens provided as input to the policy for each model. Although models like ResNet-50 and DINOv2 support both Global and Dense features, we report all results using their Dense representations, as these consistently yielded superior performance in both in-domain and out-of-domain evaluations. Additionally, our comparison includes a segmentation-driven object-centric baseline which integrates the Segment Anything Model (SAM)~\cite{kirillov2023segment} with DINOv2~\cite{oquab2024dinov2learningrobustvisual} inspired by~\cite{shi2024composingpretrainedobjectcentricrepresentations}.

\paragraph{Impact of robotic pre-training}
To understand the impact of domain alignment, Table~\ref{table:abl_data} ablates the choice of different robotic datasets used for pre-training. A mixture of diverse robotic datasets not only improves performance over using any individual robotic dataset, but also outperforms the standard vision baseline (COCO) across all evaluation domains. 
\begin{table}[h]
\centering
\small
\caption{\textbf{Pre-training data evaluation.} Final global performance for DINOSAUR-Rob*.}
\label{table:abl_data}
\resizebox{\columnwidth}{!}{%
\begin{tabular}{|l||c|c|c|}
\hline
Pre-training data  & MetaWorld & LIBERO  & Real-Robot \\
\hline
BridgeV2 (B) \cite{walke2024bridgedatav2datasetrobot} & \underline{0.75} & 0.73 & 0.45 \\
Fractal (F) \cite{brohan2023rt1roboticstransformerrealworld} & 0.72 & 0.58 & 0.38 \\
DROID (D) \cite{khazatsky2024droidlargescaleinthewildrobot} & 0.74 & 0.72 & 0.26 \\
\rowcolor{gray!10} DINOSAUR-Rob* (D+B+F) & \textbf{0.76} & \textbf{0.77} & \textbf{0.56} \\
\hline
\end{tabular}%
}
\end{table}

\begin{table*}[h!]
    \centering
    \caption{\textbf{Models overview.} Comparison of the data and sizes of the different pre-trained visual models . ViT: Vision Transformer, SOCR: Slot-based Object-Centric layers, G: Global, D: Dense, SM: Segmentation.}
    \begin{tabular}{|c||c|c|c|c|c|}
        \hline
        \textbf{Model}                                                & \textbf{Backbone} & \textbf{Pre-training Dataset}                                            & \textbf{\# of params.} & \textbf{Features} & \textbf{\# of tokens} \\
        \hline
        DINOv2 \cite{oquab2024dinov2learningrobustvisual}             & ViT               & LVD-142M \cite{oquab2024dinov2learningrobustvisual}                      & 86M                    & D/G                 & 196/1                  \\
        VC-1 \cite{majumdar2024searchartificialvisualcortex}          & ViT               & Ego4D \cite{grauman2022ego4dworld3000hours} and ImageNet \cite{russakovsky2015imagenetlargescalevisual} & 86M                    & G                 & 1                    \\
        Theia \cite{shang2024theiadistillingdiversevision}            & ViT               & ImageNet \cite{russakovsky2015imagenetlargescalevisual}                  & 140M                   & D                 & 196                  \\
        R3M \cite{nair2022r3muniversalvisualrepresentation}           & ResNet-50         & Ego4D \cite{grauman2022ego4dworld3000hours}                              & 25.6M                  & G                 & 1                    \\
        ResNet-50 \cite{he2015deepresiduallearningimage}              & ResNet-50         & ImageNet \cite{russakovsky2015imagenetlargescalevisual}                  & 25.6M                  & D/G                 & 49/1                   \\
        SAM \cite{kirillov2023segment}  + DINOv2                      & ViT + SM          & LVD-142M \cite{oquab2024dinov2learningrobustvisual}  + SA-V \cite{kirillov2023segment}                  & $\sim$400M             & SM                & 10                 \\
        \rowcolor{gray!10} DINOSAUR* & ViT + SOCR        & COCO \cite{lin2015microsoftcococommonobjects}                            & 88M                    & Slot              & 10                   \\
        \rowcolor{gray!10} DINOSAUR-Rob* & ViT + SOCR        & Robot Mixt.                                                              & 88M                    & Slot              & 10                  \\
        \hline
    \end{tabular}
    \label{tab:models}
\end{table*}


\section{Results}
\label{sec:result}
We evaluate the role of different visual features types in learning and generalizing robotic manipulation policies. Our work aim to answer the following questions: 
\begin{itemize}
    \item \textbf{Q1:} Can SOCRs improve robot policy learning efficiency over other visual representations? 
    \item \textbf{Q2:} Can SOCRs enhance policy generalization under visual distribution shifts ? 
    \item \textbf{Q3:} When do SOCRs fail, and what factors impacts their performance ? 
\end{itemize}
In simulation, we report mean success rates over three random seeds with 50 rollouts per task; in the real world, we conduct 12 rollouts per task and per generalization level ($\sim$1000 rollouts in total by combining all experiments).

\begin{figure*}[h]
	\centering
	\includegraphics[width=0.9\linewidth]{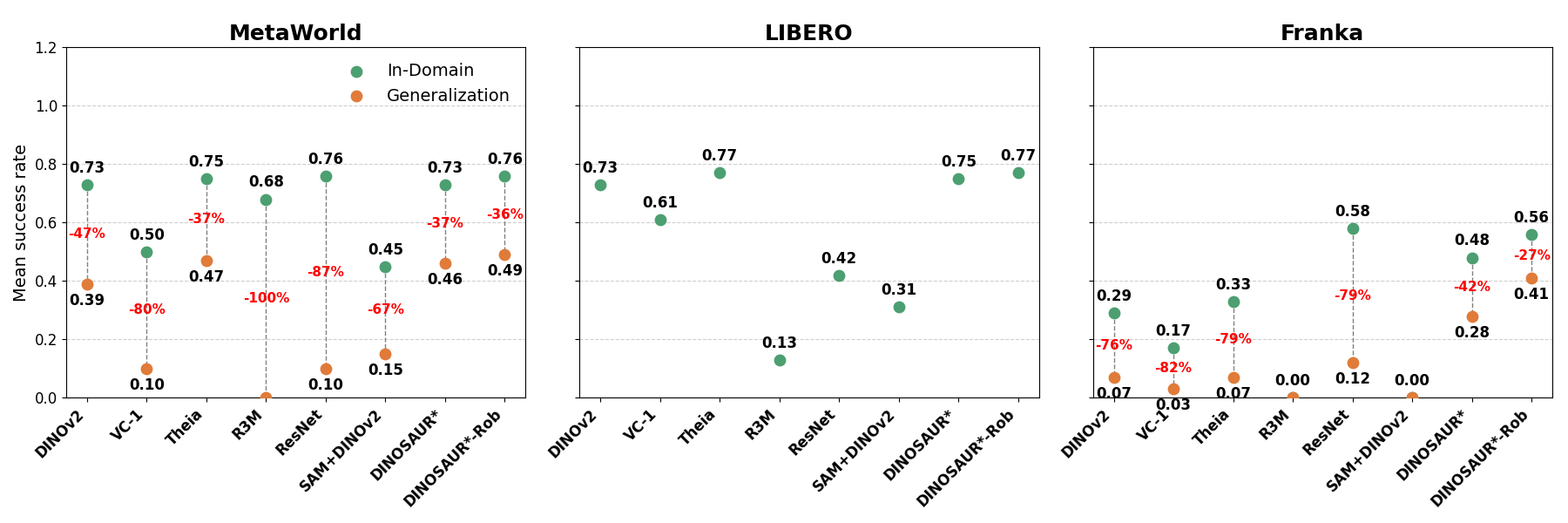}
	\caption{\textbf{Overall success rate on in-domain and generalization scenarios.} Mean success rate over all tasks for each visual model on MetaWorld (left), LIBERO (middle) and Real robot using Franka (right). 
	Green dot: in-domain performance, Orange dot: average performance over all generalization scenarios (distractors, novel textures, lighting changes). Red number: relative drop in performance from in-domain to generalization settings. As LIBERO's benchmark does not introduce any distributional shifts for evaluation, we only report in-domain performances.}
	\label{fig:perf}
\end{figure*}

\subsection{Q1: Do SOCRs improve manipulation policy learning ?}
Figure~\ref{fig:perf} summarizes the average success rate across MetaWorld, LIBERO, and our real-world benchmark. The green dots represent in-domain performance, while the orange dots indicate average success rates across various generalization scenarios (distractors, novel textures, lighting changes). The red numbers denote the relative drop in performance from in-domain to generalization settings. Policy models based on object-centric features, especially DINOSAUR-Rob*, consistently achieve the highest overall performance, on par with or surpassing the best dense and global baselines across all environments. This confirms that SOCRs not only achieve effective policy learning but also scale well to complex, multi-object scenarios.

We observe that the segmentation-driven representation (SAM+DINOv2)\cite{shi2024composingpretrainedobjectcentricrepresentations, zhu2023learninggeneralizablemanipulationpolicies} fails to learn effective control policies. To ensure a fair comparison, we restricted this baseline to encode only object appearance (the "what") without spatial coordinates. This lack of spatial grounding severely hampers the model's ability to capture structural relationships crucial for manipulation. Furthermore, relying on heuristic masking pipelines (Grounded SAM) introduces two critical limitations: first, severe temporal instability introduces massive variance that destabilizes training. Second, the heavy computational overhead creates a systemic bottleneck, introducing latency that makes reactive, closed-loop control impractical in the real world. Together, these failures highlight the distinct advantages of end-to-end learned object-centric representations (SBOCRs), which efficiently capture joint spatial-appearance features and maintain the temporal consistency required for high-frequency robotic control.

In \textbf{MetaWorld}, all models except the VC-1-based perform above 60\%. The low VC-1 performance may result from its MAE-based pretraining, which is sensitive to domain mismatch when not fine-tuned. Object-centric-based models perform comparably to top baselines here, despite the simplicity of the environment.

In \textbf{LIBERO}, which features complex scenes with multiple objects, SOCRs-based policies perform on par with or better than dense-based (Theia, DINOv2). The global-based models (R3M, VC-1) lag behind, likely due to their inability to capture fine-grained object interactions.

In the \textbf{real-world}, the ResNet-based policy is surprisingly strong, likely due to the diversity of ImageNet pretraining, consistent with \cite{dasari2023unbiasedlookdatasetsvisuomotor}. The DINOSAUR-Rob* policy performs on par with ResNet and outperforms all other baselines by a significant margin, achieving a 56\% success rate. Notably, even without robotic pretraining, the DINOSAUR* policy remains highly competitive, achieving a 48\% success rate on real-world tasks and improve 20\% over the DINOv2 policy. This suggests that the intermediate object-centric structuration itself confers robustness and effectiveness.

Overall, our results for Q1 confirm that \textbf{object-centric models are not only effective but scalable across domains}, offering performance benefits in both structured simulation tasks and noisy real-world environments.

\subsection{Q2: Do SOCRs enhance generalization under visual distribution shifts?}
We now evaluate generalization to out-of-distribution conditions, including novel distractors, unseen textures, and lighting changes. Table~\ref{table:perf_gen_meta}, and \ref{table:perf_gen_real} provide detailed results for each environment and shift type. As SAM+DINOv2 fails to learn effective policies, we exclude it from the generalization analysis. 

Overall, SOCRs-based policies exhibit substantially better robustness to distribution shifts compared to dense and global representations. Notably, DINOSAUR-Rob* policy consistently achieves the highest success rates overall across environments, with the smallest relative performance drop from in-domain to generalization settings (Figure~\ref{fig:perf}). Even without robotic pretraining, DINOSAUR* remains highly competitive in simulation and clearly outperforms all baselines in the real world. Notably, we can directly compare DINOSAUR* and DINOv2, as they share the same backbone with only an additional adapted Slot-Attention layer on top of the frozen model. This comparison highlights the benefits of object-centric intermediate representations for generalization: DINOSAUR* significantly outperforms DINOv2 across all scenarios, with the exception of the distractor setting, a specific behavior and limitation that we analyze in detail in the final section of the paper.

\begin{table}[h]
\caption{\textbf{Generalization MetaWorld.} Comparison of different levels of generalization in MetaWorld.}
\label{table:perf_gen_meta}
    \centering
    \resizebox{\columnwidth}{!}{%
        \begin{tabular}{|c|c|c|c|c|}
        \hline
            Models & Distractors  & Textures & Lighting & Over. \\
            \hline
            DINOv2 &   0.39 ± 0.04  & 0.27 ± 0.02 &  0.52 ± 0.03 & 0.39 \\
            VC1  & 0.06 ± 0.02  & 0.0 ± 0.0  & 0.23 ± 0.03 & 0.10\\
            Theia & \textbf{0.65} ± 0.10  & 0.28 ± 0.06  & 0.48 ± 0.10 & \underline{0.47}\\
            R3M & 0.0 & 0.0  & 0.0 & 0.0 \\
            ResNet-50 & 0.04 ± 0.02  & 0.0 ± 0.0 & 0.22 ± 0.02 & 0.10\\
            \rowcolor{gray!10} DINOSAUR* & 0.21 ± 0.04 & \textbf{0.48} ± 0.06 & \textbf{0.71} ± 0.06 & 0.46 \\
            \rowcolor{gray!10} DINOSAUR-Rob* & \underline{0.46 ± 0.14} & \underline{0.36 ± 0.05} & \underline{0.65 ± 0.14} & \textbf{0.49}\\
        \hline
        \end{tabular}%
    }
\end{table}

In \textbf{MetaWorld} (Table~\ref{table:perf_gen_meta}), DINOv2 and Theia-based policies outperform ResNet-based policies on average, confirming previous findings (e.g.,\cite{burns2023makespretrainedvisualrepresentations}). Notably, Theia performs best in the distractor scenario, likely due to CLIP-based text-image alignment helping the policy to ignore irrelevant patches.  Remarkably, the SOCRs-based models excel under texture and lighting shifts, where they outperform all baselines by a large margin. This suggests that object-centric representations effectively capture invariant object properties, enhancing robustness to appearance changes. Indeed, Figure~\ref{fig:overview} shows that SOCRs can distinguish objects from background clutter, likely enabling policies to focus on task-relevant elements. 

\begin{table}[h]
\caption{\textbf{Generalization Real-world.} Comparison of different levels of generalization in the real robot.}
\label{table:perf_gen_real}
	\begin{center}
		\begin{tabular}{|c|c|c|c|}
		\hline
			Models & Distractors  & Textures & Overall \\
			\hline
			DINOv2 & 0.06  & 0.08 & 0.07 \\
			VC1  & 0.03   & 0.02  & 0.03 \\
			Theia & 0.06  & 0.08  & 0.07 \\
			R3M & 0.0 & 0.0 & 0.0 \\
			ResNet-50 & 0.15 & 0.10 &  0.12 \\
			\rowcolor{gray!10} DINOSAUR* & \underline{0.27} & \underline{0.29} & \underline{0.28} \\
			\rowcolor{gray!10} DINOSAUR-Rob* & \textbf{0.37} & \textbf{0.44} & \textbf{0.41} \\
		\hline
		\end{tabular}
	\end{center}
\end{table}

In \textbf{real-world evaluations} (Table~\ref{table:perf_gen_real}), both SOCRs-based models excels in the two generalization levels and overperform by a large margin every other features types. The ResNet-50-based policy, which performed competitively in-domain, struggles significantly under distribution shifts, likely due to its limited capacity to disentangle objects from background noise. Dense-based models also see substantial performance drops, underscoring their sensitivity to visual perturbations. In contrast, SOCRs-based models maintain robust performance, with DINOSAUR-Rob* achieving a 41\% success rate on average across shifts, highlighting the practical benefits of object-centric representations for real-world robotic manipulation.

In summary, results for Q2 demonstrate that \textbf{object-centric representations generalize better across diverse distribution shifts}, particularly those that perturb low-level appearance. This robustness likely stems from SOCRs ability to filter task-irrelevant background and focus on object-level structure.

\subsection{Q3: When do SOCRs fail, and what factors impacts their performance ?}
\label{sec:abla}
\begin{figure}[t]
    \centering
    \includegraphics[width=\linewidth]{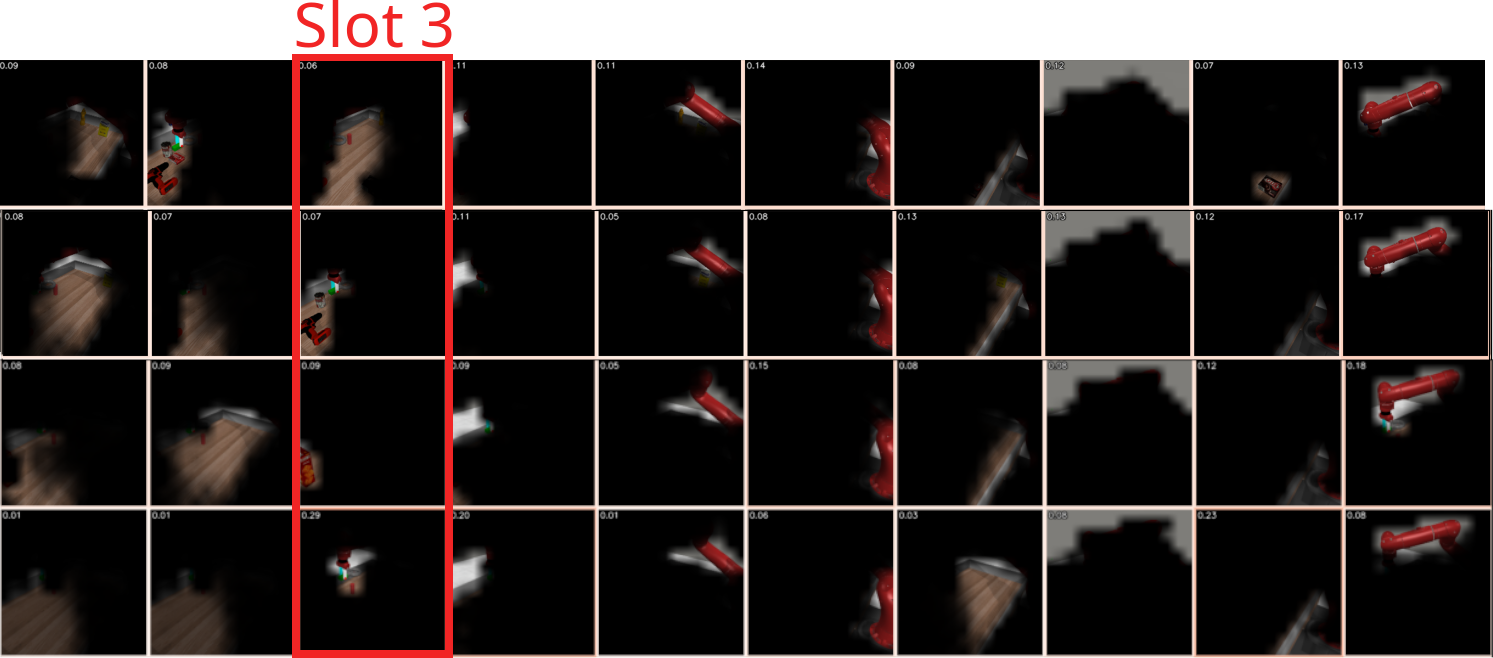}
    \caption{\textbf{Qualitative analysis of distractor generalization failures (K=10).} We visualize the slot decomposition produced by the DINOSAUR* model across four difficulty levels in MetaWorld. From top to bottom: \textit{Hard}, \textit{Medium}, and \textit{Easy} distractor scenarios, and \textit{In-Domain} baseline (bottom row). Ideally, the gripper and object (Slot 3 in the baseline) should remain isolated. However, we observe \textit{slot merging} in the different levels of clutter.}
    \label{fig:slot_abla}
\end{figure}
Even though object-centric models are more robust to distractors than most methods, there is still a significant performance drop compared to other distribution shifts. To diagnose the root cause of this performance degradation, we analyze the slot attention maps across varying levels of visual distraction (Figure~\ref{fig:slot_abla}). While the slots capturing the background and the robot arm remain largely unaffected by the shift, we identify a key phenomena that degrades representation quality: \textbf{slot merging}. This occurs when features from distractor objects leak into task-relevant slots (e.g., those containing the target object or the gripper), effectively polluting critical state information with irrelevant visual noise.

These findings indicate that standard SOCRs, constrained by a fixed-capacity bottleneck, struggle to separate novel objects from task-relevant ones when objects outnumber slots. Consequently, a critical question arises regarding mitigation: Does increasing the number of slots resolve these overcrowding effects?

\paragraph{Impact of slot number}
\begin{figure*}[t]
    \centering
    \includegraphics[width=0.75\linewidth]{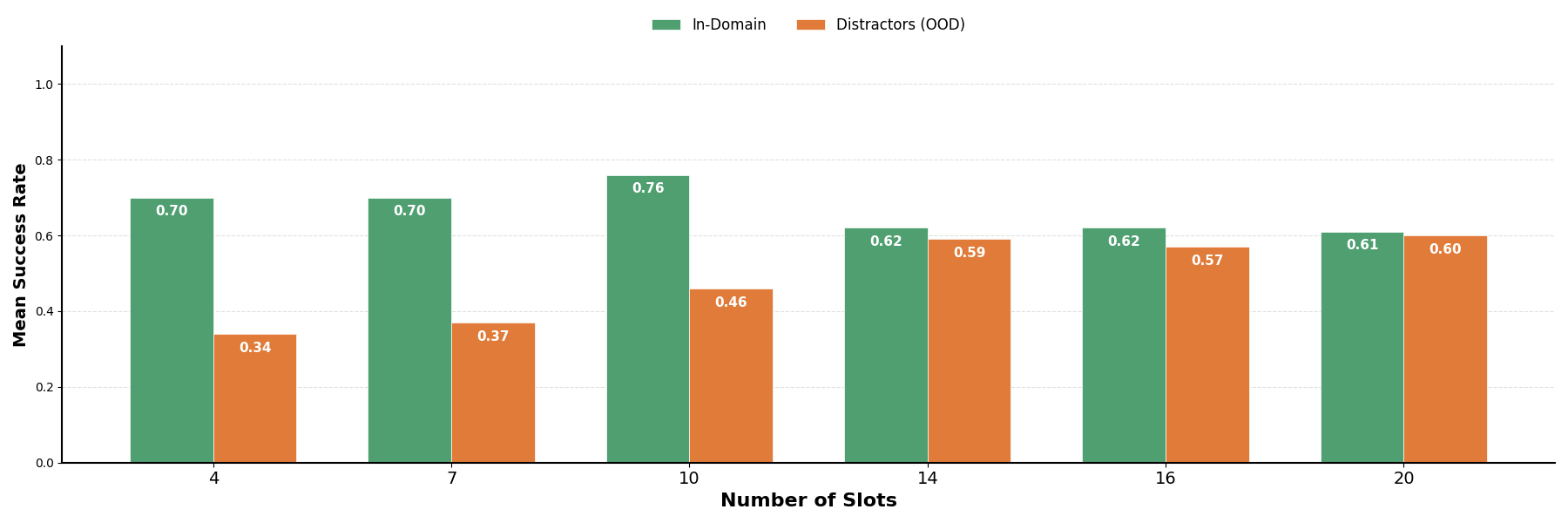}
    \caption{\textbf{Slot count ablation.} We evaluate the impact of varying the number of slots ($K$) at inference using a DINOSAUR-Rob* model trained with $K=10$. Reducing capacity ($K < 10$) exacerbates sensitivity to distractors, dropping OOD success rates significantly (to 0.34 at $K=4$) due to forced slot merging. Conversely, increasing capacity ($K > 10$) improves OOD robustness (reaching up to 0.60 at $K=20$) but causes a notable decline in In-Domain (ID) performance, highlighting a trade-off between representation stability against distractors and policy optimization complexity.}
    \label{fig:ablation_num}
\end{figure*}
A distinct advantage of SOCR models is their flexibility to modify the number of slots ($K$) dynamically at inference time, as slots are initialized from a shared learnable Gaussian distribution~\cite{locatello2020objectcentriclearningslotattention}. We leverage this property to investigate whether modulating the information bottleneck can mitigate distractor interference, and avoid \textit{slot merging}. Keeping the pre-trained DINOSAUR-Rob* model frozen, we extract representations using different slot capacities $K \in \{4, 7, 10, 14, 16, 20\}$ and train separate versions of the policy for each configuration. We evaluate these models across both In-Domain (ID) and Distractor (OOD) scenarios in MetaWorld. The results, illustrated in 

Figure~\ref{fig:ablation_num}, reveal an interesting trade-off between peak ID performance and OOD robustness depending on the slot capacity:
\begin{itemize}
    \item \textbf{Under-segmentation ($K < 10$):} We observe a sharp performance collapse when reducing capacity, particularly at $K=4$ where the OOD success rate drops to $34\%$. This confirms the "slot merging" failure mode identified previously: when $K$ is lower than the number of distinct entities in the scene (robot, target, goal, plus distractors), the model is forced to compress distractors and task-relevant objects into shared slots, rendering the state representation ambiguous for the policy.
    \item \textbf{Increased Capacity ($K > 10$):} Increasing the slot budget to $K=14, 16,$ or $20$ demonstrates a clear trade-off. While In-Domain performance decreases (dropping from $0.76$ at $K=10$ to $\sim0.61$ at higher capacities), the robustness to OOD distractors notably improves, peaking at $0.60$ for $K=20$. The extra slots allow the visual encoder to isolate novel distractors into their own representations rather than merging them with task-relevant features.
    However, when there are less objects on the scene (ID, no distractors), the extra slots might decompose relevant objects into multiple parts, which can make the representation less useful for the policy, leading to a decline in ID success rate.
\end{itemize}

In summary, higher slot budgets ensure the visual encoder has enough bandwidth to allocate individual slots to novel OOD distractors without corrupting the representations of the robot or the target. By keeping these task-relevant features isolated, the fundamental state representation remains stable, which is why OOD performance actually climbs and stabilizes around 0.60 at higher capacities. However, passing a larger set of slots to the downstream policy introduces a new optimization challenge: the policy must now learn to identify and act upon the correct subset of objects out of a noisier set of vectors that represents objects and parts. The consistent degradation of ID performance (dropping from 0.76 to roughly 0.61) suggests that the policy architecture struggles to efficiently filter through this increased dimensionality. Ultimately, these findings indicate that while providing a larger slot budget is an effective strategy for avoiding distractor interference at the representation level, it is not a complete solution. 

To address these limitations, we propose \textbf{two directions} for next-generation SOCR architectures. \textbf{Multi-granular representations}: models should utilize fine-grained slots to capture visual complexity and coarse-grained slots to aggregate semantic entities, decoupling visual noise from policy inputs. \textbf{Adaptive capacity and Semantics}: dynamic slot capacity must be paired with selection mechanisms to distinguish active novel objects from passive background noise~\cite{aydemir2023selfsupervisedobjectcentriclearningvideos, fan2024adaptiveslotattentionobject}. Furthermore, future work must bridge the gap between unsupervised grouping and semantic understanding~\cite{Didolkar_2025_CVPR}, ensuring that slots align with functional categories (e.g., "tool," "obstacle") rather than merely visual clusters.

\section{Conclusion}
We formalized and empirically validated the structural bottleneck hypothesis for visuomotor control, showing that object-level abstraction, not backbone scale alone, drives robustness under distribution shift. We show that SOCR-based policies (specifically DINOSAUR-Rob*) consistently outperform traditional dense and global baselines. These benefits are most pronounced under severe distribution shifts in lighting and texture, confirming that object-centric inductive biases better reflect the structured nature of physical interaction than pixel-level features.

However, our analysis also highlights that robustness is not automatic. We observed that while SOCRs resist appearance shifts, they remain sensitive to high-clutter scenarios. Ablation studies on slot capacity ($K$) demonstrate a fundamental trade-off between representation fidelity and policy optimization. Specifically, a low capacity ($K<10$) leads to under-segmentation, corrupting the state representation by merging distractor features into task-relevant slots. Conversely, a high capacity ($K>10$) preserves the representation by isolating distractors, but it degrades in-domain performance by flooding the downstream policy with a noisy, overloaded state space that the architecture struggles to filter.

\section*{ACKNOWLEDGMENT}
This work was in part supported by the French Research Agency, l'Agence Nationale de Recherche (ANR), through the projects Chiron (ANR-20-IADJ-0001-01), Aristotle (ANR-21-FAI1-0009-01), Astérix (ANR-23-EDIA-0002), Demeter (ANR-25-HTCE-0002) and Protheus ( ANR-25-TSIA-0011-01), the French national investment prioritary program PSPC FAIR WASTE project, the French-Singaporean collaborative Embodied AI project within the CREATE, as well as a donation to Fonds de Dotation Centrale Lyon by Huawei Technologies R\&D France. It was granted access to the HPC resources of IDRIS under the allocation 2025-[AD011016842], 2025-[AD011016615] and 2026-[A0201017513] made by GENCI.
 {
  \bibliographystyle{IEEEtran.bst}
  \bibliography{IEEEabrv.bib}
 }
 
\clearpage 
\end{document}